\documentclass{article}
\usepackage{spconf,amsmath,graphicx,hyperref}
\usepackage{arydshln}
\usepackage{amssymb}
\usepackage{booktabs}
\usepackage{multirow}

\ninept
\title{Frame-Difference Guided Dynamic Region Perception for CLIP Adaptation in Text-Video Retrieval}
\name{Jiaao Yu$^{1}$, Mingjie Han$^1$, Tao Gong$^1$, Jian Zhang$^2$, Man Lan$^{*,1}$
\thanks{$^{*}$ denotes the corresponding author.}
}
\address{
$^1$ \small School of Computer Science and Technology, East China Normal University, China \\
$^2$ \small School of Information Science and Technology, University of Science and Technology of China. China
}

\begin{document}
\maketitle

\begin{abstract}

\end{abstract}
%100-150

With the rapid growth of video data, text-video retrieval technology has become increasingly important in numerous application scenarios such as recommendation and search. Early text-video retrieval methods suffer from two critical drawbacks: first, they heavily rely on large-scale annotated video-text pairs, leading to high data acquisition costs; second, there is a significant modal gap between video and text features, which limits cross-modal alignment accuracy.
With the development of vision-language model, adapting CLIP to video tasks has attracted great attention. However, existing adaptation methods generally lack enhancement for dynamic video features and fail to effectively suppress static redundant features. To address this issue, this paper proposes FDA-CLIP (Frame Difference Alpha-CLIP), which is a concise CLIP-based training framework for text-video alignment. Specifically, the method uses frame differences to generate dynamic region masks, which are input into Alpha-CLIP as an additional Alpha channel. This proactively guides the model to focus on semantically critical dynamic regions while suppressing static background redundancy. Experiments demonstrate that frame difference-guided video semantic encoding can effectively balance retrieval efficiency and accuracy.


%\cite{tcsvt1,,tzeng2017adversarial,shen2018wasserstein,long2017deep}

\section{Introduction}
\label{sec:intro}

With the growth of video data, Text-Video Retrieval has become increasingly important in recommendation and search scenarios; early methods relying on independent training of video-text specific encoders required massive annotated data and had limited cross-modal alignment accuracy.
Constructing text-video datasets incurs high annotation costs; thus, a highly promising solution is to leverage pre-trained text-image \cite{radford2021learning,yu2023chinese,shao2023unified} foundation models and transfer their well-acquired powerful representation capabilities to the video domain. 

Early video-text retrieval \cite{chen2020fine,liu2021hit,bain2021frozen,han2021fine,wang2021t2vlad} methods had two core limitations: first, they relied on large-scale annotated video-text pairs for effective training, incurring high data acquisition costs; second, the modal gap between video and text features was significant, and independent encoding made fine-grained semantic alignment difficult, limiting retrieval accuracy. This dilemma was broken only with the emergence of CLIP-based methods: by pre-training on tens of millions of static image-text pairs, CLIP built a universal cross-modal feature space. Adapting it to video tasks enables low-cost reuse of its strong cross-modal alignment capabilities, without the need to build video encoders from scratch.

The challenge in transferring CLIP \cite{radford2021learning} from static images to dynamic video domains lies in balancing efficiency and retrieval accuracy.
The early representative work, CLIP4Clip \cite{luo2021clip4clip}, extracts features from key video frames using the CLIP image encoder, aggregates them into video-level representations via average pooling, and finally computes cosine similarity with text representations to complete retrieval. However, features of numerous static background frames and redundant frames in videos dilute the semantic information of key frames (e.g., action transition frames and object interaction frames), limiting cross-modal alignment accuracy.
Subsequent researchers improved on this issue: \cite{gorti2022x} introduced a transformer cross-attention \cite{chen2021crossvit} module to dynamically model inter-frame dependencies conditioned on text queries; \cite{liu2022ts2} designed a gated fusion unit to filter frame features. Additionally, some works adopted soft prompting methods \cite{deng2023prompt} for feature fusion.
However, previous methods focused solely on efficient feature fusion and overlooked enhancing key information during the encoding stage. Instead, we propose a novel approach that can achieve promising results even with simple average pooling for feature aggregation.

\begin{figure*}[t]
\begin{center}
   \includegraphics[width=1.0\linewidth]{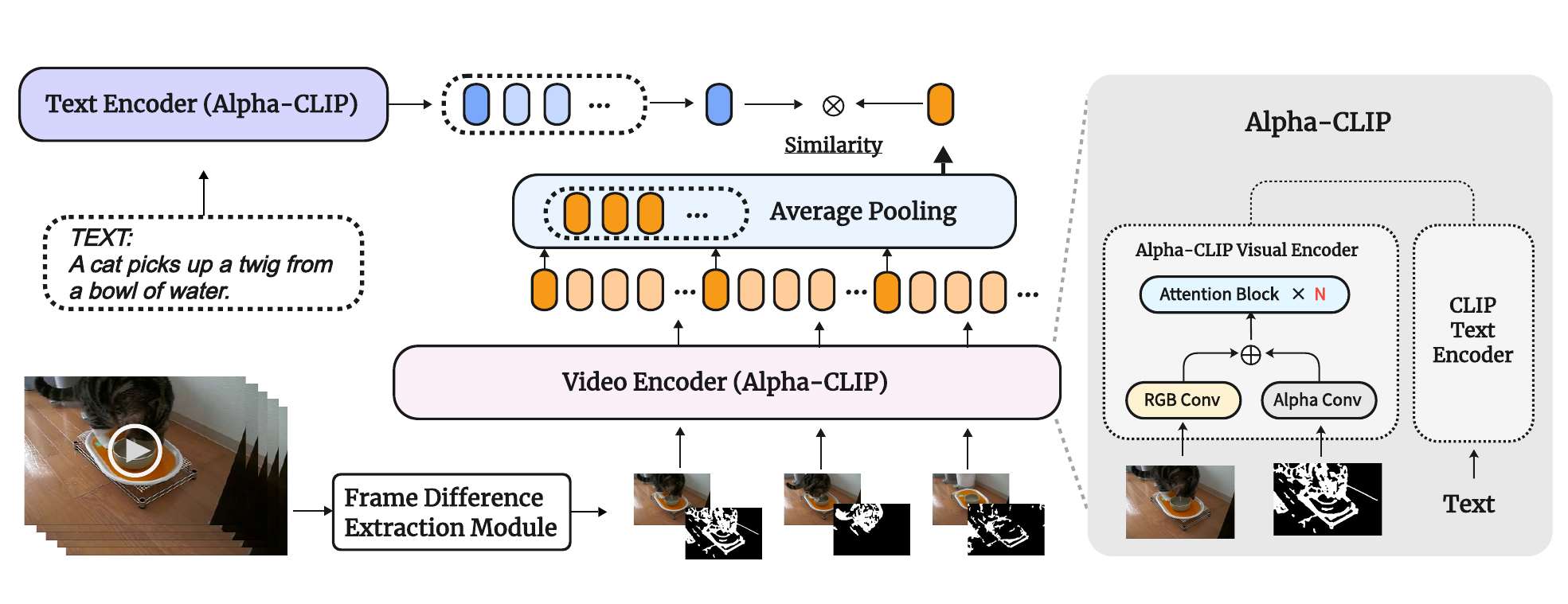}
\end{center}
   \caption{This figure illustrates the FDA-CLIP framework.}
\label{fig:main}
\end{figure*}

Previous studies on frame difference for video semantic encoding are relatively scarce, yet frame difference can accurately characterize changes in dynamic regions between frames—this is crucial for understanding the semantics of consecutive frames. Currently, Alpha-CLIP \cite{sun2024alpha}, an improved version of CLIP, enables the integration of frame difference for video semantic encoding: as shown on the right side of Figure \ref{fig:main}, Alpha-CLIP modifies CLIP’s visual encoder by adding a convolutional layer (Conv layer) for encoding single-channel masks. It fuses the mask features with RGB image features and performs incremental training, ultimately achieving enhanced understanding of information in the masked regions of images. With Alpha-CLIP, we can annotate key information in video frames, thereby assisting the visual 

This paper proposes a concise and efficient CLIP-adapted method for vision-text matching, named Frame Difference Alpha-CLIP (FDA-CLIP). Its core idea is to use frame difference information to compute masks, then leverage Alpha-CLIP to achieve dynamic region-guided cross-modal alignment.
First, Alpha-CLIP [35] introduces a learnable Alpha channel to proactively regulate the model's attention weights on local image regions, this property provides a natural advantage for focusing on dynamic regions in videos.
Based on this, we design a three-step core process: 1) Frame difference mask generation: Calculate pixel differences between adjacent key video frames, obtain dynamic region masks via adaptive threshold binarization, and filter out dynamic regions strongly related to text; 2) Alpha-CLIP encoding: Input dynamic region masks as the Alpha channel to guide Alpha-CLIP to focus only on dynamic regions, avoiding static redundancy interference; 3) Lightweight feature aggregation: Frame features can be aggregated into video representations simply through average pooling, without the need for additional cross-modal fusion modules.

\section{Methods}
\label{sec:format}

\subsection{Preliminaries}
Given a set of video clips $V$ and a set of captions $T$, the goal of this work is to learn a similarity function $s(v_i,t_j)$ that measures the similarity between a video clip and a caption. The core objective of $s(v_i,t_j)$ is to output high similarity scores for relevant video-text pairs and low scores for irrelevant pairs.

\subsection{Approach Overview}

As shown in Figure \ref{fig:main}, we propose the Frame Difference Alpha-CLIP (FDA-CLIP) framework.

This method is based on the pre-trained CLIP model, employing an image encoder with the ViT architecture. Given an input video, each video frame is first split into non-overlapping patches of fixed size, which are then linearly projected into 1D patch embeddings. After CLIP processing, the $[CLS]$ embedding is concatenated with the embedding sequence of each frame—this sequence has been pre-trained to capture local semantic features within the sequence. In Alpha-CLIP, the visual encoder not only encodes the RGB channels but also an additional Alpha channel, followed by feature fusion before inputting into the Attention block. We use the Frame Diff module to compute frame differences, which are structured into a Mask. This Mask is fed into Alpha-CLIP as the extra Alpha channel for visual feature encoding. For each frame, we extract its $[CLS]$ Token, and obtain a 1D feature vector for the entire video clip via average pooling.
For the text encoding stage, we retain the original settings of CLIP’s Text Encoder, and finally use the $[CLS]$ Token as the 1D feature vector corresponding to the text input.
The cosine similarity between the two 1D vectors is then calculated to evaluate the matching degree between the video clip and the text paragraph.

\subsection{Frame Difference Extraction}

To save computational resources, frame sampling is performed on videos. Given that all video clips in the dataset are relatively short, 6 frames are uniformly sampled during the training phase, and 12 frames during the testing phase.
The frame difference Extraction module aims to extract dynamically changing regions from consecutive video frames, providing accurate spatial attention guidance for subsequent visual encoding. Its core process is divided into two stages: frame difference calculation and mask optimization, with specific implementations as follows:
First, two adjacent frames (the previous frame and the current frame) are converted to grayscale images to simplify computation. Let the pixel values at coordinate $(x,y)$ in the grayscale images be $G_{t-1}(x,y)$ for the previous frame and $G_{t}(x,y)$ for the current frame. A frame difference map $D$ is obtained via pixel-wise absolute difference operation, which $D = |G_{t}(x,y) - G_{t-1}(x,y)|$, a larger value indicates more significant changes at the corresponding position between the two frames. 

Subsequently, the frame difference map is converted into a binary mask through thresholding, as shown in Equation \ref{eq:eq1}, where $\tau$ is a preset threshold, default is 25.

\begin{equation}
\begin{aligned}
    \label{eq:eq1}
        M(x,y) = 
        \begin{cases} 
        255 & \text{if } D(x,y) > \tau \\
        0 & \text{otherwise}
        \end{cases}
\end{aligned}
\end{equation}

To eliminate noise interference and improve the integrity of dynamic regions, multi-step post-processing is applied to the initial mask: first, morphological closing is used to fill small holes inside dynamic regions, followed by morphological opening to remove isolated noise points at region edges; then median filtering is adopted to suppress salt-and-pepper noise; finally, connected component analysis is performed to filter out tiny connected components with an area less than 50 pixels, eliminating meaningless local disturbances such as sensor noise and minor light variations.
Through the above process, this module can effectively capture dynamic regions strongly correlated with text semantics in videos (e.g., object movement, action interaction), while suppressing static backgrounds and redundant information, thus providing a reliable spatial prior for the attention regulation of Alpha-CLIP.

% MSVD

\begin{table*}[htbp]
\centering
\caption{Performance comparison of different methods on MSVD, CLIP4Clip and CenterCLIP have multiple versions in their
papers, here we choose the versions with the highest Meta Sum values.}
\label{tab:1}
\scalebox{1.01}{
\begin{tabular}{lccccccccc}
\toprule
\multirow{2}{*}{\textbf{Methods}} & \multicolumn{4}{c}{\textbf{Text $\Rightarrow$ Video}} & \multicolumn{4}{c}{\textbf{Video $\Rightarrow$ Text}} & \multirow{2}{*}{\textbf{Meta Sum }$\uparrow$} \\
\cmidrule(lr){2-5} \cmidrule(lr){6-9}
 & \textbf{R@1} $\uparrow$ & \textbf{R@5} $\uparrow$ & \textbf{R@10} $\uparrow$ & \textbf{MnR} $\downarrow$ & \textbf{R@1} $\uparrow$ & \textbf{R@5} $\uparrow$ & \textbf{R@10} $\uparrow$ & \textbf{MnR} $\downarrow$ & \\
\midrule
CLIP2TV \cite{gao2021clip2tv} & 47.0 & 76.5 & 85.1 & 10.1 & - & - & - & - & - \\
CLIP2Video \cite{fang2021clip2video} & 47.0 & 76.8 & 85.9 & 9.6 & 58.7 & 85.6 & 91.6 & 4.3 & 445.6 \\
X-CLIP \cite{ma2022x} & 47.1 & \textbf{77.8} & - & 9.5 & 60.9 & 87.8 & - & 4.7 & - \\
X-Pool \cite{gorti2022x} & 47.2 & 77.4 & 86.0 & 9.3 & 66.4 & 90.0 & 94.2 & 3.3 & 461.2 \\
\midrule
CLIP4Clip \cite{luo2022clip4clip} & 45.2 & 75.5 & 84.3 & 10.3 & 62.0 & 87.3 & 92.6 & 4.3 & 446.9 \\
CenterCLIP \cite{zhao2022centerclip} & 47.4 & 76.5 & 85.2 & 9.7 & 62.7 & 88.1 & 92.8 & 4.1 & 452.7 \\
Prompt Switch \cite{deng2023prompt} & 47.1 & 76.9 & \textbf{86.1} & \textbf{9.5} & 68.5 & 91.8 & \textbf{95.6} & \textbf{2.8} & 466.0 \\
$Ours$ & \textbf{48.2} & 77.3 & 85.8 & 9.8 & \textbf{70.2} & \textbf{92.8} & 95.5 & \textbf{2.8} & \textbf{469.8} \\
\bottomrule
\end{tabular}
}
\end{table*}

\begin{figure*}[!htbp]
\begin{center}
   \includegraphics[width=0.901\linewidth]{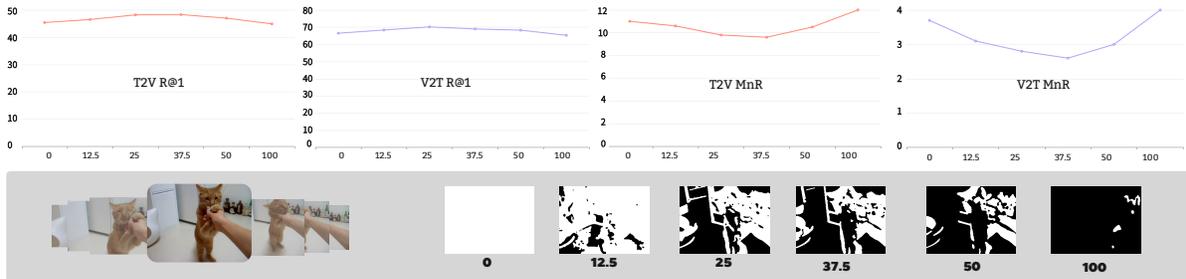}
\end{center}
   \caption{Sensitivity Verification of Hyperparameter $\tau$.}
\label{fig:vis}
\end{figure*}

\subsection{Loss Function}

We use a symmetric cross entropy loss $\mathcal{L}$ over these similarity scores to train the model’s parameters as Equation \ref{eq:eq2}.

\begin{equation}
\begin{aligned}
    \label{eq:eq2}
        \begin{aligned}
        \mathcal{L}_{v2t} &= -\frac{1}{B} \sum_{i=1}^{B} \log \frac{\exp(s(v_i, t_i))}{\sum_{j=1}^{B} \exp(s(v_i, t_j))}, \\
        \mathcal{L}_{t2v} &= -\frac{1}{B} \sum_{i=1}^{B} \log \frac{\exp(s(v_i, t_i))}{\sum_{j=1}^{B} \exp(s(v_j, t_i))}, \\
        \mathcal{L} &= \frac{1}{2}(\mathcal{L}_{v2t} + \mathcal{L}_{t2v}).
        \end{aligned}
\end{aligned}
\end{equation}

\section{Experiment}
\label{sec:experiment}

\subsection{Datasets and Implementation details}

We validate our model on MSR-VTT \cite{2016MSR} and MSVD \cite{chen:acl11}. MVSD comprises 1,970 videos with durations spanning from 1 to 62 seconds, and each video is paired with approximately 40 English sentences. MSR-VTT contains 10,000 videos, each with a length that ranges from 10 to 32 seconds and 200,000 captions. For both datasets, we employed a learning rate of $1e-6$.

To evaluate our model’s performance, we adopt three standard retrieval metrics: Recall@K (R@K, higher values indicate better performance), Median Rank (MdR, lower values are preferable), and Mean Rank (MnR, lower values are preferable). Specifically, Recall@K (R@K) quantifies the proportion of test samples where the correct target (e.g., the matching video for a text query, or the corresponding text for a video query) is among the top-K results retrieved for a given query sample.

\begin{table*}[htbp]
\centering
\caption{Performance comparison of different methods on MSRVTT.}
\label{tab:2}
\scalebox{1.0}{
\begin{tabular}{lccccccccc}
\toprule
\multirow{2}{*}{\textbf{Methods}} & \multicolumn{4}{c}{\textbf{Text $\Rightarrow$ Video}} & \multicolumn{4}{c}{\textbf{Video $\Rightarrow$ Text}} & \multirow{2}{*}{\textbf{Meta Sum $\uparrow$}} \\
\cmidrule(lr){2-5} \cmidrule(lr){6-9}
 & \textbf{R@1 $\uparrow$} & \textbf{R@5 $\uparrow$} & \textbf{R@10 $\uparrow$} & \textbf{MnR $\downarrow$} & \textbf{R@1 $\uparrow$} & \textbf{R@5 $\uparrow$} & \textbf{R@10 $\uparrow$} & \textbf{MnR $\downarrow$} & \\
\toprule
% \multicolumn{10}{l}{\textit{cross-modal temporal fusion}} \\
CLIP2TV \cite{gao2021clip2tv} & 46.1 & 72.5 & 82.9 & 15.2 & 43.9 & 73.0 & 82.8 & 11.1 & 401.2 \\
CLIP2Video \cite{fang2021clip2video} & 45.6 & 72.6 & 81.7 & 14.6 & 43.3 & 72.3 & 82.1 & 10.2 & 397.6 \\
EMCL \cite{jin2022expectation} & 46.8 & 73.1 & 83.1 & - & 46.5 & 73.5 & 83.5 & - & 406.5 \\
X-CLIP \cite{ma2022x} & 46.1 & 73.0 & 83.1 & 13.2 & \textbf{46.8} & 73.3 & 84.0 & 9.1 & 406.3 \\
DRL \cite{wang2022disentangled} & 47.4 & 74.6 & 83.8 & - & 45.3 & 73.9 & 83.3 & - & \textbf{408.3} \\
X-Pool \cite{gorti2022x} & \textbf{46.9} & 72.8 & 82.2 & 14.3 & 44.4 & 73.3 & 84.0 & 9.0 & 403.6 \\
\midrule
% \multicolumn{10}{l}{\textit{text-agnostic temporal pooling}} \\
CLIP4Clip \cite{luo2022clip4clip} & 44.5 & 71.4 & 81.6 & 15.3 & 42.7 & 70.9 & 80.6 & 11.6 & 391.7 \\
CenterCLIP \cite{zhao2022centerclip} & 44.2 & 71.6 & 82.1 & 15.1 & 42.8 & 71.7 & 82.2 & 11.1 & 394.6 \\
TS2-Net \cite{liu2022ts2} & 44.4 & 72.1 & 82.2 & 14.6 & 43.7 & 70.8 & 80.4 & 11.6 & 393.6 \\
Prompt Switch \cite{deng2023prompt} & 46.1 & 72.8 & 81.8 & 14.4 & 44.8 & 73.7 & 82.4 & 9.9 & 401.6 \\
$Ours$ & 45.4 & \textbf{73.1} & \textbf{82.6} & \textbf{14.1} & 45.8 & \textbf{74.4} & \textbf{84.4} & \textbf{8.7} & 405.7 \\
\bottomrule
\end{tabular}
}
\end{table*}

\subsection{Performance Comparison}

To verify the effectiveness of the proposed method, we conducted large-scale comparative experiments, with comparison against two categories of baseline methods: cross-modal feature aggregation methods (e.g., CLIP2TV \cite{gao2021clip2tv}, CLIP2Video \cite{fang2021clip2video}, EMCL \cite{jin2022expectation}, X-CLIP \cite{ma2022x}, DRL \cite{wang2022disentangled}, X-Pool \cite{gorti2022x}) and other mainstream methods (e.g., CLIP4Clip \cite{luo2022clip4clip}, CenterCLIP \cite{zhao2022centerclip}, TS2-Net \cite{liu2022ts2}, Prompt Switch \cite{deng2023prompt}). Experimental results show that the proposed method, which employs average pooling for feature aggregation, outperforms all baselines that use the same aggregation strategy. More notably, it also outperforms most cross-modal feature aggregation baselines and achieves leading performance across multiple evaluation metrics, fully demonstrating the superiority of the proposed method.

Table \ref{tab:1} presents the performance comparison results of the proposed FDA-CLIP method and two categories of baseline methods on the MSVD dataset. The results show that FDA-CLIP exhibits significant advantages in both text to video (Text→Video) and video to text (Video→Text) cross-modal retrieval tasks:
In the text to video retrieval subtask, FDA-CLIP achieves a R@1 score of 48.2\%, which is not only higher than that of baseline methods using the same average pooling aggregation strategy but also outperforms cross-modal feature aggregation methods such as CLIP2TV, CLIP2Video, and X-CLIP, with only a slightly lower score than CenterCLIP. Even without introducing complex cross-modal fusion modules, it still maintains retrieval coverage comparable to mainstream methods, and its Mean Rank (MnR) score is roughly on par with CenterCLIP demonstrating that the dynamic region-guided encoding strategy can effectively suppress redundant information interference.
In the video to text retrieval subtask, FDA-CLIP shows a more remarkable performance improvement: its R@1 score of 70.2\% significantly outperforms all baseline methods, representing a 1.7 percentage point increase compared to the second-best method (Prompt Switch). Both its R@5 and R@10 scores reach state-of-the-art levels, and its MnR score is consistent with that of Prompt Switch, making it the only method in this task direction that balances high recall and low ranking error.
In terms of the comprehensive performance metric Meta Sum, FDA-CLIP ranks first among all methods with a score of 469.8. This confirms that under the guidance of frame differences, more effective video semantic encoding is achieved.

Table \ref{tab:2} presents the performance comparison results of the proposed FDA-CLIP method and methods on the MSR-VTT dataset.
In the text to video retrieval subtask, FDA-CLIP still significantly outperforms baseline methods that adopt the same average pooling strategy. Its better Mean Rank (MnR) performance indicates that our method effectively reduces ranking errors, which proves that the frame difference-guided dynamic region perception strategy can still accurately suppress the interference of static redundant information on semantic encoding even in the large-scale data scenario of MSR-VTT.
In the video to text retrieval subtask: FDA-CLIP’s R@1 metric is higher than that of methods such as CLIP4Clip and CenterCLIP, and its R@5 metric outperforms Prompt Switch. Notably, its MnR metric is the best among all compared methods, which is 0.3 and 1.2 lower than that of X-Pool and Prompt Switch, respectively—fully demonstrating the method’s advantage of low ranking errors in retrieval.
FDA-CLIP shows a more prominent advantage in the video to text subtask, which also indirectly verifies the effectiveness of frame difference as prior information to guide the model in video semantic encoding.

\subsection{Hyperparameter Sensitivity Ablation Study}

To further verify the effectiveness of the proposed method, we conducted comparisons with baseline methods under identical experimental configurations as shown in Table \ref{tab:3}, and the proposed method achieved significant improvements across all evaluation metrics.

As shown in Fig. 2, we conduct a sensitivity verification experiment on the hyperparameter $\tau$ using the MSVD dataset, presenting R@1 results for the text to video and video to text subtasks. The results demonstrate: our method exhibits hyperparameter stability, with the optimal $\tau$ value being 25.
When $\tau = 0$, the mask is fully white (equivalent to baseline CLIP, treating all pixel regions as dynamic); as $\tau$ increases, the dynamically perceived region shrinks and feature focus improves (e.g., at $\tau = 25$, white areas in the mask represent the kitten’s contour and actions); when $\tau = 100$, the mask is nearly fully black, and performance decreases accordingly. Notably, although our method is robust to hyperparameters, the optimal $\tau$ value still needs experimental determination in practical applications.

\begin{table}[htbp]
\normalsize
\centering
\caption{Comparison with the original baseline on MSVD.\\}
% \vspace{-2mm}
\renewcommand{\arraystretch}{0.9}
\scalebox{1.1}{
\begin{tabular}{lccccc}
\cline{1-6}
\multicolumn{2}{c}{\textbf{Video $\Rightarrow$ Text}} & \textbf{R@1} & \textbf{R@5} & \textbf{R@10} & \textbf{MnR}\\
\cline{1-6}
% $\times$
  \multicolumn{2}{c}{CLIP-Based} & 66.5 & 89.2 & 93.5 & 3.7\\
  \multicolumn{2}{c}{FDA-CLIP(ours)}  & 70.2 & 92.8 & 95.5 & 2.8\\
\cline{1-6}
\end{tabular}}
\label{tab:3}
\end{table}

\section{Conclusion}

This paper proposes a lightweight framework named FDA-CLIP, which introduces a frame difference-guided dynamic region perception mechanism into the Alpha-CLIP architecture. The core innovation of FDA-CLIP lies in: generating dynamic region masks via frame difference operations, and feeding them into the model as an additional Alpha channel to proactively guide the model to focus on semantically critical regions while suppressing static background redundancy.
Experimental results on the MSVD and MSR-VTT datasets verify the effectiveness of the proposed method: FDA-CLIP outperforms all baseline models using the same average pooling strategy, shows prominent advantages in the video-to-text subtask, and exhibits strong stability against hyperparameter variations. These results confirm that: taking inter-frame differences as prior information can significantly improve video semantic encoding without relying on complex cross-modal fusion modules.
Future work will focus on two aspects: first, extending FDA-CLIP to long video sequences; second, integrating temporal modeling capabilities to better understand dynamic events. Such extensions are expected to further enhance the efficiency and robustness of video-text retrieval systems in real-world scenarios.
Code can be found: \url{https://github.com/yjainqdc/FDACLIP}

\vfill\pagebreak

\bibliographystyle{IEEEbib}
\bibliography{main}

\begin{thebibliography}{10}

\bibitem{radford2021learning}
Alec Radford, Jong~Wook Kim, Chris Hallacy, Aditya Ramesh, Gabriel Goh,
  Sandhini Agarwal, Girish Sastry, Amanda Askell, Pamela Mishkin, Jack Clark,
  et~al.,
\newblock ``Learning transferable visual models from natural language
  supervision,''
\newblock in {\em International conference on machine learning}. PmLR, 2021,
  pp. 8748--8763.

\bibitem{yu2023chinese}
Haiyang Yu, Xiaocong Wang, Bin Li, and Xiangyang Xue,
\newblock ``Chinese text recognition with a pre-trained clip-like model through
  image-ids aligning,''
\newblock in {\em Proceedings of the IEEE/CVF International Conference on
  Computer Vision}, 2023, pp. 11943--11952.

\bibitem{shao2023unified}
Zhiyin Shao, Xinyu Zhang, Changxing Ding, Jian Wang, and Jingdong Wang,
\newblock ``Unified pre-training with pseudo texts for text-to-image person
  re-identification,''
\newblock in {\em Proceedings of the IEEE/CVF international conference on
  computer vision}, 2023, pp. 11174--11184.

\bibitem{chen2020fine}
Shizhe Chen, Yida Zhao, Qin Jin, and Qi~Wu,
\newblock ``Fine-grained video-text retrieval with hierarchical graph
  reasoning,''
\newblock in {\em Proceedings of the IEEE/CVF conference on computer vision and
  pattern recognition}, 2020, pp. 10638--10647.

\bibitem{liu2021hit}
Song Liu, Haoqi Fan, Shengsheng Qian, Yiru Chen, Wenkui Ding, and Zhongyuan
  Wang,
\newblock ``Hit: Hierarchical transformer with momentum contrast for video-text
  retrieval,''
\newblock in {\em Proceedings of the IEEE/CVF international conference on
  computer vision}, 2021, pp. 11915--11925.

\bibitem{bain2021frozen}
Max Bain, Arsha Nagrani, G{\"u}l Varol, and Andrew Zisserman,
\newblock ``Frozen in time: A joint video and image encoder for end-to-end
  retrieval,''
\newblock in {\em Proceedings of the IEEE/CVF international conference on
  computer vision}, 2021, pp. 1728--1738.

\bibitem{han2021fine}
Ning Han, Jingjing Chen, Guangyi Xiao, Hao Zhang, Yawen Zeng, and Hao Chen,
\newblock ``Fine-grained cross-modal alignment network for text-video
  retrieval,''
\newblock in {\em Proceedings of the 29th ACM International Conference on
  Multimedia}, 2021, pp. 3826--3834.

\bibitem{wang2021t2vlad}
Xiaohan Wang, Linchao Zhu, and Yi~Yang,
\newblock ``T2vlad: global-local sequence alignment for text-video retrieval,''
\newblock in {\em Proceedings of the IEEE/CVF conference on computer vision and
  pattern recognition}, 2021, pp. 5079--5088.

\bibitem{luo2021clip4clip}
Huaishao Luo, Lei Ji, Ming Zhong, Yang Chen, Wen Lei, Nan Duan, and Tianrui Li,
\newblock ``Clip4clip: An empirical study of clip for end to end video clip
  retrieval,''
\newblock {\em arXiv preprint arXiv:2104.08860}, 2021.

\bibitem{gorti2022x}
Satya~Krishna Gorti, No{\"e}l Vouitsis, Junwei Ma, Keyvan Golestan, Maksims
  Volkovs, Animesh Garg, and Guangwei Yu,
\newblock ``X-pool: Cross-modal language-video attention for text-video
  retrieval,''
\newblock in {\em Proceedings of the IEEE/CVF conference on computer vision and
  pattern recognition}, 2022, pp. 5006--5015.

\bibitem{chen2021crossvit}
Chun-Fu~Richard Chen, Quanfu Fan, and Rameswar Panda,
\newblock ``Crossvit: Cross-attention multi-scale vision transformer for image
  classification,''
\newblock in {\em Proceedings of the IEEE/CVF international conference on
  computer vision}, 2021, pp. 357--366.

\bibitem{liu2022ts2}
Yuqi Liu, Pengfei Xiong, Luhui Xu, Shengming Cao, and Qin Jin,
\newblock ``Ts2-net: Token shift and selection transformer for text-video
  retrieval,''
\newblock in {\em European conference on computer vision}. Springer, 2022, pp.
  319--335.

\bibitem{deng2023prompt}
Chaorui Deng, Qi~Chen, Pengda Qin, Da~Chen, and Qi~Wu,
\newblock ``Prompt switch: Efficient clip adaptation for text-video
  retrieval,''
\newblock in {\em Proceedings of the IEEE/CVF International Conference on
  Computer Vision}, 2023, pp. 15648--15658.

\bibitem{sun2024alpha}
Zeyi Sun, Ye~Fang, Tong Wu, Pan Zhang, Yuhang Zang, Shu Kong, Yuanjun Xiong,
  Dahua Lin, and Jiaqi Wang,
\newblock ``Alpha-clip: A clip model focusing on wherever you want,''
\newblock in {\em Proceedings of the IEEE/CVF conference on computer vision and
  pattern recognition}, 2024, pp. 13019--13029.

\bibitem{gao2021clip2tv}
Zijian Gao, Jingyu Liu, Weiqi Sun, Sheng Chen, Dedan Chang, and Lili Zhao,
\newblock ``Clip2tv: Align, match and distill for video-text retrieval,''
\newblock {\em arXiv preprint arXiv:2111.05610}, 2021.

\bibitem{fang2021clip2video}
Han Fang, Pengfei Xiong, Luhui Xu, and Yu~Chen,
\newblock ``Clip2video: Mastering video-text retrieval via image clip,''
\newblock {\em arXiv preprint arXiv:2106.11097}, 2021.

\bibitem{ma2022x}
Yiwei Ma, Guohai Xu, Xiaoshuai Sun, Ming Yan, Ji~Zhang, and Rongrong Ji,
\newblock ``X-clip: End-to-end multi-grained contrastive learning for
  video-text retrieval,''
\newblock in {\em Proceedings of the 30th ACM international conference on
  multimedia}, 2022, pp. 638--647.

\bibitem{luo2022clip4clip}
Huaishao Luo, Lei Ji, Ming Zhong, Yang Chen, Wen Lei, Nan Duan, and Tianrui Li,
\newblock ``Clip4clip: An empirical study of clip for end to end video clip
  retrieval and captioning,''
\newblock {\em Neurocomputing}, vol. 508, pp. 293--304, 2022.

\bibitem{zhao2022centerclip}
Shuai Zhao, Linchao Zhu, Xiaohan Wang, and Yi~Yang,
\newblock ``Centerclip: Token clustering for efficient text-video retrieval,''
\newblock in {\em Proceedings of the 45th international ACM SIGIR conference on
  research and development in information retrieval}, 2022, pp. 970--981.

\bibitem{2016MSR}
Jun Xu, Tao Mei, Ting Yao, and Yong Rui,
\newblock ``Msr-vtt: A large video description dataset for bridging video and
  language,''
\newblock in {\em Conference on Computer Vision and Pattern Recognition
  (CVPR)}, 2016.

\bibitem{chen:acl11}
David~L. Chen and William~B. Dolan,
\newblock ``Collecting highly parallel data for paraphrase evaluation,''
\newblock in {\em Proceedings of the 49th Annual Meeting of the Association for
  Computational Linguistics (ACL-2011)}, Portland, OR, June 2011.

\bibitem{jin2022expectation}
Peng Jin, Jinfa Huang, Fenglin Liu, Xian Wu, Shen Ge, Guoli Song, David
  Clifton, and Jie Chen,
\newblock ``Expectation-maximization contrastive learning for compact
  video-and-language representations,''
\newblock {\em Advances in neural information processing systems}, vol. 35, pp.
  30291--30306, 2022.

\bibitem{wang2022disentangled}
Qiang Wang, Yanhao Zhang, Yun Zheng, Pan Pan, and Xian-Sheng Hua,
\newblock ``Disentangled representation learning for text-video retrieval,''
\newblock {\em arXiv preprint arXiv:2203.07111}, 2022.

\end{thebibliography}

\end{document}